# Applying Deep Learning to Ads Conversion Prediction in Last-mile Delivery Marketplace

Di Li*
DoorDash Engineering
United States
di.li2@doordash.com

Xiaochang Miao*
DoorDash Engineering
United States
xiaochang.miao@doordash.com

Huiyu Song
DoorDash Engineering
United States
heather.song@doordash.com

Chao Chu
DoorDash Engineering
United States
chao.chu@doordash.com

Hao Xu
DoorDash Engineering
United States
hao.xu@doordash.com

Mandar Rahurkar
DoorDash Engineering
United States
mandar.rahurkar@doordash.com

## ABSTRACT

Deep neural networks (DNNs) have revolutionized web-scale ranking systems, enabling breakthroughs in capturing complex user behaviors and driving performance gains. At DoorDash, we first harnessed this transformative power by transitioning our homepage Ads ranking system from traditional tree-based models to cutting-edge multi-task DNNs. This evolution sparked advancements in data foundations, model design, training efficiency, evaluation rigor, and online serving, delivering substantial business impact and reshaping our approach to machine learning.

In this paper, we talk about our problem-driven journey—from identifying the right problems and crafting targeted solutions, to overcoming the complexity of developing and scaling a deep learning recommendation system. Through our successes and learned lessons, we aim to share insights and practical guidance to teams pursuing similar advancements in machine learning systems.

*These authors contributed equally to this work.

## 1 INTRODUCTION

DoorDash Ads aims to empower local restaurants to grow their businesses by connecting consumers with more diverse yet relevant dining options. To serve the best interests of merchants and advertisers, DoorDash Ads employs a Cost Per Acquisition (CPA) pricing model. Thus, advertisers are charged only when a consumer places an order. This design makes post-view conversion rate (CVR) prediction the cornerstone of the real-time auction. More specifically, accurate CVR predictions are essential to prioritize high-quality ads, ensuring a fair auction that benefits both advertisers and consumers.

Traditional tree-based models provide a solid foundation [10, 11, 19] but fall short in capturing the complexity and diversity of consumer interactions, especially as DoorDash scales. Deep neural networks (DNNs) overcome these limitations through advanced representational learning, enabling them to process large-scale, heterogeneous data such as temporal trends [41], contextual signals [30], and textual [8], visual [13, 39], and graphical features [38].

Additionally, DNNs open new opportunities for cross-domain knowledge sharing through transfer learning and enable holistic user behavior modeling via multi-task learning (MTL). This is especially important for DoorDash Ads, a nascent business vertical with limited data. MTL allows the model to leverage insights from user personalization on organic surfaces while also promotes synergies across related tasks throughout the entire shopping journey [24, 33], such as predicting click-through rate (pCTR), add-to-cart rate (pATC), and conversion rate (pCVR).

Transitioning to deep neural networks (DNNs) at DoorDash's scale is far more than a simple shift in model architecture—it is a transformative overhaul of the entire system. This journey was undertaken through a series of iterative advancements, each addressing new challenges and uncovering opportunities for innovation. It involves a comprehensive transformation of the offline model training, calibration, troubleshooting, and online serving infrastructure.

Reflecting on both our successes and failures, this work shares hard-earned lessons with those embarking on a similar mission, with the hope of making their journey a bit smoother.

The major contributions of this paper include:

- **Domain-Driven Model Design:** Tailored **M**ulti-**T**ask **M**ulti-**L**abel (MTML) architectures and feature engineering that address the unique challenges of Ads ranking. Our work demonstrates how domain-specific ML design can effectively balance exploration and exploitation, offering a replicable framework for similar challenges.
- **Hybrid Embedding Strategies:** Leveraging cutting-edge LLM-based pre-trained embeddings integrated with domain-specific ID-based features. This novelty achieves a balance between generalization and specification, unlocking areas of improvements in both exploration and exploitation.
- **A Framework to Bridge Online-Offline Gaps:** Introducing a framework that leverages offline replay of data generation to diagnose issues such as feature staleness and cached residuals. This powerful instrumentation uncovers domain-specific problems while offering a practical and scalable approach that can be adopted to address similar challenges.
- **Real-Time Serving Optimization:** Delivering low-latency predictions through the integration of the PyTorch 2.0 [3], FBGEMM [15], and a re-architected Ad Exchange (AdEx) service. These advancements enable the scalable deployment of more complex DNN models while maintaining high responsiveness of the entire Ads serving system.

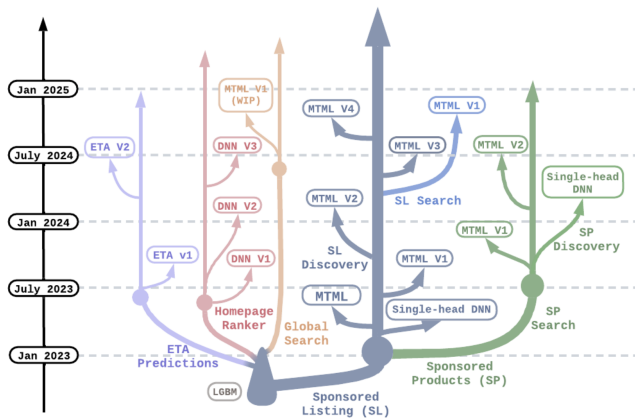

Figure 1: The evolution of DNN / MTML models at DoorDash, first from Ads then to broader applications.

Our advancements in DNN technologies and system design have significantly influenced other product ML teams within DoorDash. By sharing our learnings, methodologies, and platform capabilities, we have enabled these teams to leverage the power of DNNs for their specific applications, unlocking success across various domains. The evolutionary roadmap of DNN models, as illustrated in Figure-1, reflects this journey—from initial adoption in restaurant ads (i.e. sponsored listings) to widespread applications across key business verticals, including delivery ETA predictions, home-feed ranking, global search, and consumer packaged goods (CPG) ads (i.e. sponsored product).

The rest of this paper is organized as follows. In Section 2, we review related work to contextualize our approach within the broader landscape of deep learning advancements. In Section 3, we describe the system architecture underpinning DoorDash Ads, focusing on the integration of offline pipelines and online serving workflows. Section 4 talks about our model's feature categories, in which we highlight the most predictive features. In Section 5 and 6, we discuss our training and evaluation methodologies, including distributed frameworks and model evaluation metrics. Section 7 delves into our DNN model evolution in which we adopt a thoughtful problem-driven innovation by identifying and tackling key challenges encountered. Section 8 covers our framework for diagnosing and bridging the online-offline model performance gap, and in section 9 we unveil our real-time serving optimizations. Finally, in Section 10, we conclude by summarizing the key takeaways from this transformation and outlining future directions for deep learning applications in ad ranking systems.

## 2 RELATED WORK

Deep neural networks (DNNs) are renowned for their ability to learn complex, non-linear relationships and patterns among features, reducing the need for extensive manual feature engineering while providing the flexibility to design tailored model architectures for real-world problems. It has been a trend and proven to be successful in ranking related scenarios in past decades. Google designed Deep & Wide [7] for recommendation system ranking aiming at achieving both generalization and memorization; Airbnb migrated their search model from Tree-based model to DNN after some failed experiences [10]; Youtube Video Recommendations [8] automatically learn user embeddings and video embeddings to tackle the difficulty of large, sparse and noisy business scenario; Deep interest evolution network (DIEN)[41] explicitly models user behavior sequence and their temporal interests; DCN [35] and DCN-V2 [36] are designed to learn complicated feature crossing for large-scale online advertising industry.

Furthermore, MTML architectures allow share of feature representations, leading to the optimization for the entire shopping journey through transfer learning while achieving overall lower training and serving cost. Therefore it has been adopted by more and more use cases, such as feed Ads [23], Airbnb Search [33], customer acquisition [37]. To further address the challenge of data sampling bias, ESMM [25] is proposed to learn deeper funnel (e.g. CVR) conditionally. And to further improve the efficiency and specificity of sharing feature representations among tasks, advanced frameworks like Multi-gate Mixture-of-Experts (MMoE)[22] explicitly capture task relationships, enabling better coordination between objectives, which has been proved to be helpful for food recommendation[43], video recommendations[40], e-commerce[17].

Inspired by past successful experiences, we decide to make the transaction through a series of iterations, targeting for a better understanding of our customers to further improve consumers' experiences. This paper records our successes and failures to share learning for others.

## 3 SYSTEM ARCHITECTURE

The system architecture for DoorDash Ads integrates offline model training with real-time online inference, creating a cohesive end-to-end pipeline to support ad ranking and auctions as shown in Figure-2(b). For this discussion, we focus on sponsored listings presented on discovery surfaces in the home feed (see Figure-1(a) as an example).

When a consumer interacts with the DoorDash marketplace, such as visiting the homepage, the Feed Service communicates with the Ad Exchange Service. The Ad Exchange Service fetches relevant entity IDs and real-time features based on request metadata. Concurrently, online user engagement signals are processed and stored via an Apache Flink [14] job, with a Data Quality Monitoring (DQM) Service ensuring data accuracy and fidelity before persistence.

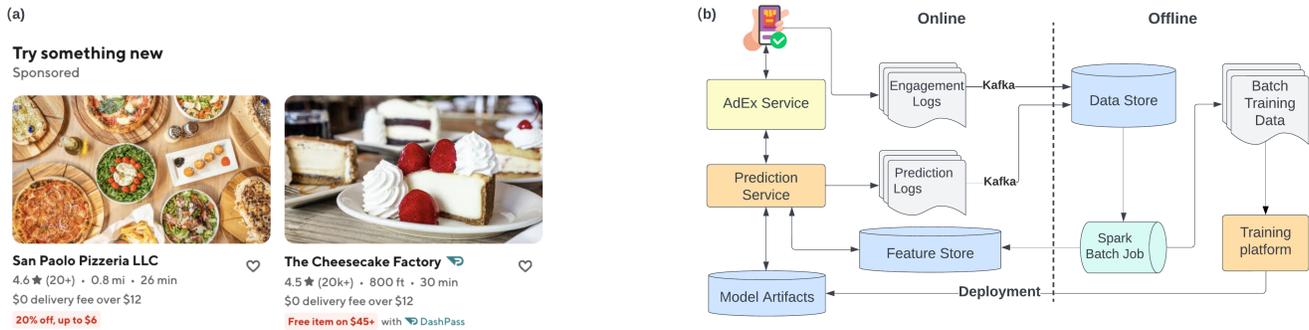

Figure 2: a) Sponsored restaurants on the DoorDash homepage. b) The overview of DoorDash Ads workflow.

The Ad Exchange (AdEx) Service sends gRPC requests to the ML Prediction Service, specifying the required model configuration. The Prediction Service processes the request and returns the predicted values, which are then used by AdEx to rank ads in the downstream auction process. To ensure continuous improvement, all features and predictions are logged for future model calibration, performance monitoring, and model retraining.

Model training is powered by a distributed framework built on Fireworks AI, leveraging PyTorch 2.0 and TorchRec for efficient Distributed Data Parallel (DDP) training across GPUs.

## 4 FEATURE CATEGORIES

In this section we list all the feature categories of our DNN model and highlight the most important features within each category.

***Dense Features***. As noted in the Representation Layer of Figure-3, the most important ones are the historical engagement features. For CVR prediction, these include aggregation over historical interactions at the user level, store level, and cross-interactions between users and stores.

***Categorical features***. The original feature values are mapped to 1-based contiguous positive integers where 0 is reserved for missing values. During training, an embedding table will be learned where each integer is bound to a unique embedding vector. The most important categorical features for our DNN models are: hour_bucket (0-23), day_of_week (1-7), and platform (ios, android, web) which are noted as *User Embeddings*, and merchant_cohort (enterprise or small-business) which is noted as *Store Embeddings*.

***Pre-trained embeddings***. Pre-trained embeddings consist of fixed-value embedding vectors where the embedding dimension is homogeneous and each element is a floating point. For our prediction task the most important pre-trained embedding features are **content-based multimodal** embeddings. We generate these embeddings for each store item, where both *Users* and *Stores* can be represented as a list of the underlying item embeddings.

***Sparse features***. They are a variable-length list of integers where each element represents a specific type of entity ID. In practice, we represent *Users* with e.g. 1) a list of store IDs that the user has previously engaged with, and 2) a list of word tokens that the user has previously searched for. On the candidate side, it is either a single-element list that contains the candidate store ID, or a list of keyword tokens that the candidate store is associated with. Since the feature has a variable length per record, we use TorchRec JaggedTensor [1] for the representation in order to maximize the training efficiency and memory utilization. Similar to Categorical features, the integer representations are mapped to learnable embedding vectors underpinning the *User* and *Store Embeddings*.

## 5 MODEL TRAINING
### 5.1 Feature pre-processing

The performance of Neural Networks has a non-trivial dependency on the distribution of input feature values, which intrinsically have distinct mean, variance, and skewness. It has been demonstrated previously that a more unified distribution across features can minimize "internal covariate shift" [12], and lead to better model performance. This also resonates with the philosophy behind the common practice of applying normalization layers before activation units in standard neural network building blocks.

To achieve a high training throughput, our design principle accounts for the fact that the data pre-processing does not involve any heavy-lifting neural network operations such as matrix multiplications, and the pre-processing is only required once per dataset. Therefore, we select to offload this workflow to large scale CPU pools given that CPU hosts are much cheaper than the GPU ones (see Table-4 in Appendix-D for a reference on the cost benchmark).

The feature pre-processing adopts the native PyTorch distributed-data-parallel (DDP) framework [18] and involves the following three steps: 1) A small amount of data is read by data loaders where key statistics like average, variation, and median are calculated; 2) the workers start over to go through all the data, applying a transformation (e.g. log scaling for power-law distributions [10]) to each chunk; and 3) the processed batch data is asynchronously saved back to storage, and will be re-loaded during all subsequent model training processes.

## 5.2 Distributed Model Training with TorchRec

The model training framework adopts the industry standard paradigm of "Single program, Multiple data" (SPMD) [21], where multiple GPU hosts can be utilized to collaboratively execute the training task. Our framework was also built on top of the TorchRec distributed training pipeline to maximize the overlap between computation and communication.

Our hybrid strategy of separating the pre-processing and model training maximizes the overall training efficiency by minimizing the GPU idle time, as data loading and processing dominate the end-to-end training time. We also build customized model serialization workflows (see Appendix-B) to accommodate efficient offline model evaluations as well as the consistency between training and serving.

## 6 MODEL EVALUATION

While the model is typically trained with the data from most recent months, evaluation is done on the subsequent two-weeks at the end of the training data time range. We build GPU based offline batch inference pipeline to facilitate model performance benchmark upon training.

The key metrics for our offline evaluations are AUC and Normalized Binary Cross Entropy (BCE) loss.

*AUC.* We pick AUC instead of AU-PRC (Area-Under-Precision-Recall-Curve) due to the fact that AUC describes the likelihood of a randomly picked positive example having a higher prediction value than a randomly picked negative example. This metric definition aligns with our goal of measuring ranking quality.

*Normalized BCE.* Equation-1 shows the mathematical definition of this metric, where $y_i \in \{0, 1\}$ and $\bar{p} = \frac{1}{N}\sum_{i=1}^{N} y_i$. Unlike AUC, this metric captures event level prediction accuracy. The normalization term makes the metric robust by being baseline-agnostic.

$$\text{Norm-BCE} = \frac{\frac{1}{N}\sum_{i=1}^{N}[y_i * log(p(x_i)) + (1-y_i) * log(1-p(x_i))]}{\bar{p} * log(\bar{p}) + (1-\bar{p}) * log(1-\bar{p})} \quad (1)$$

## 7 MODEL EVOLUTION

In this section, we demonstrate the evolution of our DNN modeling framework together with a problem-driven feature engineering approach, in which we adopt a phased progression with a thoughtful understanding of the success and improvement areas of each model iteration. Figure-3 depicts our DNN model architecture. We modularize critical model components for efficient iterations over different design choices of each component.

*Training Objective.* The multi-task layer in our DNN model architecture consolidates the training loss from individual prediction head, and in our use case each head corresponds to a binary classification task. We expose the task weights on the client side to allow tuning. Equation-2 shows the overall training loss function, where $K$ is the number of tasks and $N$ is the number of data samples. In practice, we found that neither tuning the task weights ($w^k$) nor applying task loss balancing [20] helps with the performance of our primary prediction head, although it does affect the training convergence velocity regarding the number of epochs required to achieve equivalent evaluation AUC on the primary head.

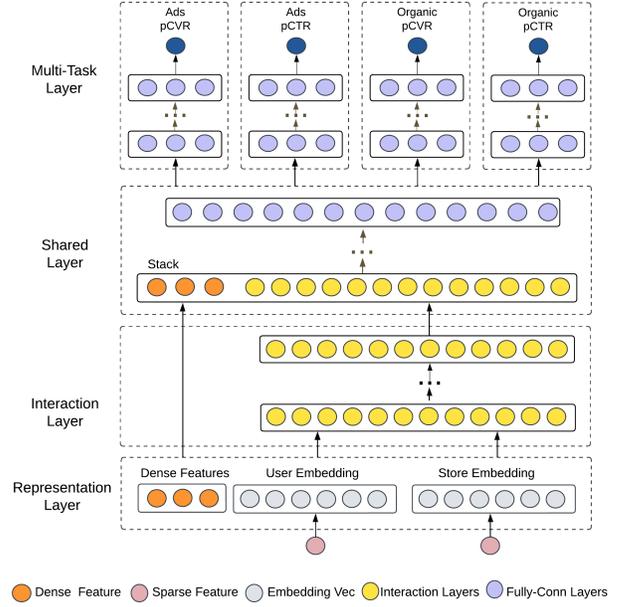

Figure 3: Our DNN model architecture.

$$\text{Loss} = \frac{1}{KN}\sum_{k=1}^{K}\sum_{i=1}^{N} w^k \cdot [y_i^k \cdot log(p^k(x_i)) + (1-y_i^k) \cdot log(1-p^k(x_i))]a \quad (2)$$

### 7.1 Milestone 1: Adopting DLRM

The journey starts from the model architecture shift from GBDT (LGBM) to DNN (Milestone 1-a), where we began with simple multi-layer perceptrons and aimed to at least achieve equivalent predictive performance as the LGBM model has gone through a few iterations and was already a strong baseline. The success in the online A/B experiment (see Row 1 in Table-2) set a solid foundation for our DNN migration.

After the successful transition, we started to explore feature engineering and model architecture opportunities that were exclusively supported by the Deep Learning Recommendation Models (DLRM) [28] framework on TorchRec, such as

- **Single head -> Multi-task learning** The limited available ads training data led to our intuitive adoption of a learning-transfer framework (see the top Multi-Task Layer in Figure-3). We first added the ads *pCTR* head, which by itself achieved +0.5% AUC gain despite the fact that it only augmented our training labels without increasing the overall training data size.
- **Introduction of sparse features** As DLRM supports modeling through variable-length user shopping history, we introduced sparse ID-based list features and started with simple mean or sum pooling method in the Interaction Layer of Figure-3. We note that due to data sparsity it is less ideal to represent users with unique consumer IDs than a list of user's past shopping history.

Although we did achieve metric wins by including the auxiliary prediction head and sparse features (Milestone 1-b), subsequent investment on the sparse ID-based features failed to payoff. As we dove deep, we identified severe model overfitting from sparse features. This echos with our expectation that due to the long-tailed distribution of store traffic and low ad adoption rate on Door-Dash, many stores have a limited number of consumer interactions with the ad. To address this, we aim to integrate organic $pCVR$ and $pCTR$ into the current MTML architecture, improving ID feature data coverage and enhancing embedding learning. Similar approaches, adopted by other teams at DoorDash later on, in scenarios with abundant training data have demonstrated significant improvements in sparse feature learning capacity while minimizing overfitting.

Furthermore, we explored alternative model architectures particularly in the *Interaction Layer* (see Figure-3). Motivated by the desired to explore widely adopted DNN architectures from industry and literature, we experimented with PNN[31] and DCN[35, 36]. However, these attempts yielded negligible performance gains, indicating that the model's performance was more constrained by the lack of high-quality **predictive features** than by **model capacity**.

In response, we redirected our focus toward extensive feature engineering in Milestone 1-c. These included features derived from the food knowledge graph for entity tagging and additional sparse list features, such as the sequence of food tags searched by the user, to construct a more comprehensive view of their shopping history.

## 7.2 Milestone 2: Identifying Model Performance Gap

As we continued to add more features, we observed a growing online vs. offline performance gap, driven primarily by feature disparities between the two environments. At this point, we began leveraging online logged features for model training, which improved alignment with online serving data. However, newly engineered features were still joined using offline pipelines. While we addressed part of the gap in this milestone, the root causes were only fully diagnosed and resolved till Milestone 3 through deeper investigation.

Although logged features helped reduce the online vs offline performance gap, the development velocity was limited by the need to collect warm-up data for newly engineered features. Hence, we proposed a hybrid data generation approach: leveraging backfilled data (offline joins) for offline model training with new features and fine-tuning with logged features prior to the deployment. This approach was necessary as the infrastructure for continuous DNN training was not yet in place.

Furthermore, a key challenge in enhancing prediction accuracy lies in better capturing user behavior for personalization. To address this, we explored several approaches to refine how user interactions with ad stores are modeled, such as adopt collective learning chain rules $pCVR = pCTR \cdot pCTCVR$ from ESMM [25]. We summarized the leanings from each attempt below, which ultimately pointing toward the need for richer personalization features.

| Event category | Fraction of positive samples | Fraction of negative samples | Online AUC |
|---|---|---|---|
| Repurchase | 67.6% | 34.3% | 0.724 |
| Exploration | 32.4% | 65.7% | 0.707 |
| Overall | 100% | 100% | 0.787 |

**Table 1: User behavior by Repurchase vs. Exploration.**

*Trial 1: Tuning negative sampling strategies*. Repurchase and exploration are the two primary user shopping behaviors on DoorDash, each with distinct conversion rates, hence different positive-to-negative sampling ratios as shown in Table-1. To balance out that, we experimented with stratified sampling ratios for negative instances. However, offline AUCs showed minimal improvement, suggesting the model's limitations were due to insufficient information to better capture user behavior with stores.

*Trial 2: Adding negative samples from dark traffic*. To expand training data and ID feature coverage, we incorporated randomly sampled "dark traffic", consisting of ad stores that participated in auctions but failed to win (i.e., no impressions). These instances are previously excluded from the training dataset.

This approach showed no significant improvement, as dark traffic primarily consists of easy negatives with low prediction values, adding little incremental value. Further analysis revealed the true bottlenecks as hard negatives (non-purchased stores with past conversions) and hard positives (unexpected conversions from exploration), contributing to 75% and 80% of top false positives and false negatives respectively.

AUC performance comparisons in Table-1 further confirms these challenges: both repurchase and exploration AUCs were lower than the overall AUC. This indicates that while the model effectively distinguishes between repurchase and exploration cohorts, it struggles to accurately predict outcomes within each group. To address these challenges, more targeted and meaningful features, along with improved model designs, are needed to enhance predictive power.

*Trial 3: Deep Interest Network (DIN) with ID-based embeddings*. To capture nuanced consumer interactions with ad stores, we switched the Interaction Layer with DIN [42]. In this approach, users were represented by a list of previously purchased and clicked store IDs, with a self-attention mechanism applied to learn deeper personalization across stores. However, this method marginally outperformed our vanilla MTML model, which aggregates sparse features using simple sum pooling.

The key limitation was the low overlap (30% on average) between ad stores and users' engaged store lists due to limited ad store adoption. Restricting the store list to only ad stores improved specificity but risked overfitting and missed broader picture of consumer preferences. This again echos the need for a four-head architecture as highlighted in Milestone 1, incorporating organic conversion and $pCTR$ predictions with a shared bottom layer to improve both store and user representational learning.

### 7.3 Milestone 3: Deep Personalization

Building on the insights from Milestone 2, we shifted our focus to tackling the nuanced challenges of deep personalization. Two key patterns emerged in the online food delivery platform: (1) users often return to order what they've purchased before, leading to high repurchase rates, and (2) users are generally cautious about exploring new options. These observations became the foundation for our next iteration.

***Identify hard negatives among Repurchases***. Conventional feature engineering approaches mostly emphasize on building predictive features that represent user's past purchase behaviors. However, the model still lacks a deep understanding of the nuanced repurchase patterns. Overall 3% data are hard negatives, where 75% of them are repurchases.

***Improve model generalization for Explorations***. Since the training data volume is highly limited for exploration events, we need innovative design of features and model architectures in order to uncover the deep personalization opportunities for generalized food preferences.

|  | AUC | CVR | Core ML improvements |
| --- | --- | --- | --- |
| Milestone 1-a | +0.5% | +3.5% | LGBM -> DNN |
| Milestone 1-b | +2.0% | +4.1% | 1) Added consumer <> hourOfDay and sparse list features<br>2) Single head -> MTML<br>3) CPU -> DDP GPU training<br>4) Standardized pre-processing |
| Milestone 1-c | +1.0% | +2.9% | Added food knowledge graph features and more sparse list features |
| Milestone 2 | +0.8% | +0.5% | 1) Used online feature logs for training data construction<br>2) Applied ESMM [25] chain rule |
| Milestone 3 | +0.6% | +1.6% | 1) Deep personalization features<br>2) Root cause and address online & offline model performance gap |
| Milestone 4 | +0.6% | +1.2% | Added store item content-based pre-trained embeddings from LLM |

Table 2: The milestones of Deep Learning progress for DoorDash Ads conversion prediction. All metrics are based on top of the previous model milestone.

Hard negatives in our case often arise from the challenge of distinguishing between broadly applicable patterns and those tied to a user's specific purchasing behaviors. For instance, 38% of false positives occurred during a less frequent ordering period (e.g., dinner) rather than the user's usual repurchase window (e.g., lunch) at a particular store. More specifically, users exhibited highly structured repurchase habits—for example, consistently ordering from a coffee shop in the morning but not at dinner. However, when users opened the app at dinner time, the model still predicted a high *pCVR* for that coffee shop, despite the user's historical behavior suggesting otherwise.

To address this, we introduced features capturing <consumer, daypart> repurchase behaviors, enabling the model to better account for nuanced preferences in repurchase patterns. Looking ahead, we plan to incorporate temporal encoding into store embeddings (inspired by positional encoding) [30], and leverage a multi-head attention transformer model to further enhance the model's ability to capture these patterns.

To suppress hard positives, our focus shifted toward achieving better generalization by capturing a broader spectrum of consumer preferences for stores, dishes, and overall price sensitivities. To this end, we enhanced our food knowledge-based features by incorporating both cuisine-level and dish-level preferences. Additionally, we integrated users' most recent search histories, store metadata (e.g., names and descriptions), and coupon engagement data (i.e., price sensitivity) to build a more comprehensive profile of user preferences across various stores.

These enhancements have yielded 0.6% AUC gain. However, to fully capitalize on the online CVR gains as reflected in Table-2, it was imperative to bridge the gap between online and offline model performance. To this end, we conducted a comprehensive investigation into this discrepancy, with a detailed analysis provided in Section 9.

### 7.4 Milestone 4: Content-based Pre-trained Embeddings

Another takeaway from Milestone 3 is that when the model lacks information to capture consumers' general and broader food interests, it leads to false negatives and jeopardizes exploration opportunities as dense features such as consumers' past engagement with the stores are usually missing for such scenario.

Previously, our efforts to improve generalization relied on directly training ID-based features and learning embeddings jointly with prediction objectives. While this approach captured some nuances, it struggled with cold start and data sparsity (e.g. ID feature overlap between user and store). An alternative is to leverage pre-trained embeddings before evolving to more complex interaction layers. In Milestone 4, we initiated this approach using content-based embeddings generated by large language models (LLMs) enriched with world knowledge.

Pre-trained embeddings address cold-start challenges by incorporating extensive external knowledge. Techniques include (a) extracting features from an encoding layer in transformer models and (b) utilizing text-to-embedding LLMs, where "text" spans multimodal inputs like item titles, descriptions, images, ratings, and prices. Table-5 in Appendix-E illustrates the trade-offs between content-based and sparse ID-based embeddings.

To balance quality and cost of serving, we selected the "text-embedding-3-large" model from OpenAI and customized the embedding dimension to 256. This allowed the integration of 6 million unique item embeddings, resulting in a model size of 3GB, nearing the capacity of our serving infrastructure. Applications with more stringent model size limitations can benefit from the built-in embedding dimension reduction technique [16, 29] by pruning the trailing elements from the embedding vectors with minimal loss of performance. Despite this constraint, even simple weighted sum pooling of pre-trained embeddings improved generalization, achieving an online AUC gain of +0.6%, and significant business metrics gains including merchant trial rate (+0.18%).

Looking ahead, pre-trained embeddings provide opportunities to enhance interaction layers. Inspired by prior work [44], we plan to model users and candidate stores as sequences of item embeddings, using dual Deep Interest Networks with pairwise cross-attention to better capture the nuanced user-candidate interactions.

## 8 ONLINE & OFFLINE MODEL PERFORMANCE

We observed a relative **4.3%** discrepancy between online and offline AUC during Milestone 3. Two factors contributed to this gap. First, the training data was three months old by the time the experiment was rolled out online. Second, Milestone 3 incorporated a substantial number of new features that lacked corresponding online logging.

To diagnose the underlying causes, we adopted a hypothesis-driven approach [9] and, through a detailed examination of the data, identified two primary drivers responsible for the observed feature disparity. Our final solution reduced the online-offline AUC gap to 0.76% in the latest iteration of our Ads Ranking model.

### 8.1 Methodology for Diagnosing Performance Discrepancies

Our initial hypotheses include:

- **Feature Generation Disparity:** This issue often arises when the offline feature pipeline fails to accurately replicate the online environment, leading to discrepancies in real-time predictions.
- **Data Distribution Shift (Concept Drift):** Changes in the underlying data distribution over time, such as seasonal variations in consumer behavior, can significantly affect model generalization.
- **Model Serving Instability:** Problems related to model serving infrastructure, such as increased latency or deployment of incorrect model versions, may also degrade online performance.

To validate these hypotheses, we reconstructed the evaluation datasets using both online logging (features from real-time serving logs) and offline joining (features joined from one day prior to impression instances) as shown in Table-3. We evaluated the model performance for each dataset to understand where the AUC gap came from.

The offline-joined datasets from the experiment period (September, Dataset 3: +2.45%) and the model development period (June, Dataset 5: +2.50%) showed similar relative AUC changes, suggesting that data distribution remained stable over time. However, the relative AUC gap between online-logged (Dataset 2: -1.80%) and offline-joined features (Dataset 3: +2.45%) during the experiment period (September) showed a significant difference of **4.25%**, which matched to the online vs offline gap we initially observed, indicating that the disparity in feature generation is the main culprit.

| Data range | Model | Feature source in eval data | Dataset index | Relative AUC |
|---|---|---|---|---|
| 3Experiment (September) | Baseline | Online Log | Dataset 1 | Baseline |
| | New model | Online Log | Dataset 2 | -1.80% |
| | New model | Offline Join | Dataset 3 | **+2.45%** |
| 2Model Dev (June) | Baseline | Offline Join | Dataset 4 | +0.77% |
| | New model | Offline Join | Dataset 5 | **+2.50%** |

Table 3: Model performance benchmark. The Baseline (production model) is trained on logged features, while the New model is trained on offline joined features.

### 8.2 Feature Staleness & Cached Residuals

To further pinpoint the root causes of this disparity, we conducted a deeper investigation and identified two key contributing factors:

*Feature Staleness*. This occurs when most recent feature values are not available during serving. Typically, this is due to delays in the feature pipeline, which may take from 1 to 2 days.

*Cached Residuals*. This occurs as the online feature upload only overrides existing keys or adds new keys without evicting the old ones in a timely manner. Features that are mostly affected are likely to have these characteristics: a) high cardinality and volatile values day-over-day, and b) aggregated over short time windows.

### 8.3 Solutions to Address the Performance Gap

We employ two complementary strategies to mitigate feature disparity:

- **Adjust Feature Join Offsets:** We generate training datasets with varying feature join offsets (e.g., -2d, -3d, -4d) and align each feature with the offset that best matches the online serving delay. This approach offers an immediate reduction in AUC discrepancies caused by *Feature Staleness*, but does not fully resolve issues related to *Cached Residuals*.
- **Enable Online Logging for Key Features:** To address *Cached Residuals*, we enable online logging for features most susceptible to residual caching. While this improves data freshness and model accuracy, it introduces trade-offs with development velocity and infrastructure complexity. Careful planning is required to balance these factors, as online logging demands additional system resources to handle increased traffic.

By combining these strategies, we can significantly narrow the performance gap and harness the CVR gains online (see Table-2).

## 9 ONLINE SERVING

Efficient online serving is critical for a high-performance ads ranking system, especially when handling real-time predictions for large-scale auctions. To meet these demands, we implemented several key optimizations across both the client-side (AdEx) and server-side (Prediction Service). Below, we detail the improvements made to reduce latency and enhance the overall efficiency of our system.

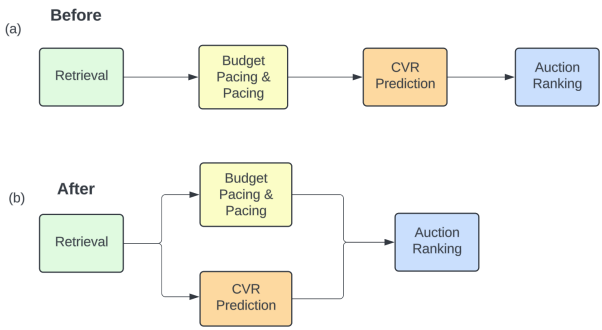

Figure 4: The transition from sequential (a) to parallelized (b) bidding and CVR prediction reduced Ad Exchange end-to-end latency by 150ms, enabling deployment of more advanced DNN models.

### 9.1 Upleveling the Inference Stack

To enhance server-side performance, we collaborated with our internal ML platform team to upgrade the prediction service to PyTorch 2.0 and integrate FBGEMM, a high-performance kernel optimized for CPU inference. This upgrade significantly reduced the latency of DNN model inference on AWS CPU instances, enabling faster and more efficient predictions.

The improved serving stack now provides a scalable and robust infrastructure for real-time production workloads at DoorDash. Its success has led to the adoption of the DNN model in the company by other ML application teams using DNN models.

### 9.2 Parallelizing Bidding and CVR Prediction

Previously, AdEx processed bidding and CVR prediction sequentially as manifested in Figure-4(a), with budget pacing & throttling and automated bidding occurring before CVR predictions. While this architecture limited the number of candidates sent to the prediction service, it constrained the latency budget for CVR prediction and thus restricted our ability to deploy more advanced ML models.

To address this, we re-architected AdEx to parallelize bidding and CVR prediction as shown in Figure-4(b). This optimization has significantly reduced end-to-end p999 latency by ~150ms. This optimization freed up additional latency budget for CVR predictions and enabled the deployment of more complex but accurate DNN models, which enhanced the precision of the ads ranking. However, this change increased the load on the prediction service by ~10%, as more store candidates bypass budget pacing before the CVR predictions. While this resulted in higher infrastructure costs, the reduced latency and improved ranking quality delivered substantial business value, making the trade-off well worth the investment.

## 10 CONCLUSION

This paper outlines our transition from tree-based models to deep learning for ads conversion prediction in the last-mile delivery marketplace. By building a scalable distributed training framework and addressing online-offline performance gaps, we significantly improved model accuracy and business outcomes. Key challenges like feature staleness and cached residuals were resolved through targeted feature engineering and online logging. Our multi-task learning architecture and hybrid embedding strategies further enhanced personalization and model generalization. These advancements not only optimized ad ranking performance but also laid a foundation for future ML innovations at DoorDash.

## 11 ACKNOWLEDGMENTS

We appreciate the partnership with Benny Chen, Pawel Garbacki, Yijie Bei, Chenyu Zhao, Dmytro Dzhulgakov, and Lin Qiao from Fireworks AI for building the foundation of the high-performance distributed training platform and all the insightful discussions on every aspect of model training, performance optimization, and platform reliability. We also want to acknowledge the remarkable internal support from the team and the company, with special thanks to Qi Chen, Igor Nodelman, Hai Yu and Shahrooz Ansari from Ads engineering organization for the collaborations and support; Kornel Csernai, Hebo Yang, Mark Yang, Ferras Hamad, Vasily Vlasov and Arvind Kalyan from Machine Learning Platform for all the collaborations on Ads User Engagement Prediction Modeling.

## A DISTRIBUTED MODEL TRAINING FRAMEWORK

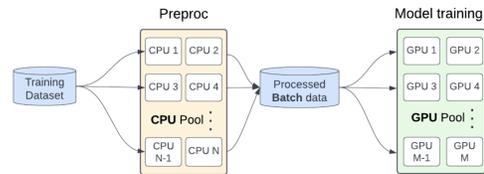

**Figure 5: The distributed model training framework.**

## B DISTRIBUTED EVALUATION AND MODEL SERIALIZATION

To ensure consistency, the feature pre-processing module for offline evaluation and online inference needs to be exactly the same as training. On the other hand, the pre-processing workflow only needs to take place once for offline evaluation, where the pre-processed data can be cached and reused for multi-epoch model snapshot evaluations. With both requirements taken into account, we spinoff the following two separate workflows:

*The deployment path.* For the purpose of online serving we generate the complete model artifact with both pre-processing and neural network layers. We use TorchScript [34] to trace and script the PyTorch model into a portable format that can be loaded and executed in environments without Python dependencies, together with various inference performance optimizations.

*The offline eval path.* This is the incomplete version of the model artifact where the preproc module is absent during model scripting. For offline multi-epoch model evaluation we take the following steps in order to maximize the efficiency:

**Step 1** Run the pre-processing job once to generate processed data for the evaluation dataset. The feature statistics (e.g. mean, median, variance) used in the pre-processing are exactly the same as the training data generation process.

**Step 2** Run the distributed inference pipeline for each model snapshot where the pre-processing module is absent. This is to ensure that the end-to-end inference workflow is consistent across both training and evaluation.

## C  FEATURE IMPORTANCE ALGORITHMS

To interpret the importance of features for DNN models, we select from three feature importance algorithms, including Integrated Gradients [32], Feature Ablation [6], and Feature Permutation [26].

We evaluate each method by retraining separate models using the top ranked $X$ ($X \in \{15, 35, 55, 75, 100\}$) features determined by each feature importance algorithm, followed by computing AUC as a function of $X$. Compared with the baseline AUC where all features are present, the closer the delta AUC curve is to line $y = 0$, the more effectively the algorithm is to capture the most predictive features.

In Figure-6, it is clear that Feature Permutation ranks the most informative features higher than other algorithms. Additionally, we consistently observed that sequence features are ranked higher in Feature Permutation compared with other algorithms. One hypothesis is that because no baseline value or feature masking is required for Feature Permutation, it more directly measures the performance drop when a single feature is shuffled.

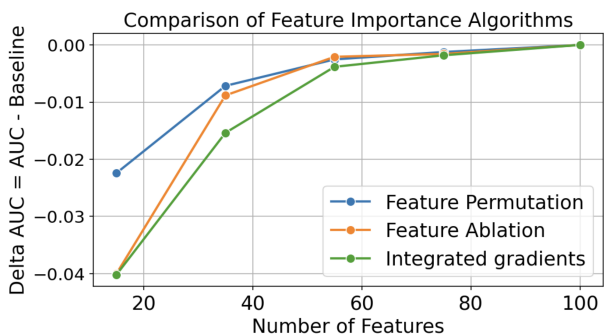

Figure 6: Evaluation AUC of Models trained with selected top X features from different Feature Importance Algorithms.

## D  CPU VS GPU HOST COST BENCHMARK

| Instance type | GPUs | vCPUs | Memory (GiB) | On-Demand Price/hr |
|---|---|---|---|---|
| t3.2xlarge |  | 8 | 32 | $0.334 |
| g4dn.12xlarge | 4 | 48 | 192 | $3.912 |

Table 4: The product details of Amazon EC2 a) low-cost general purpose CPU instances [5], and b) cost-effective GPU instances which feature NVIDIA T4 GPUs [4].

## E  PRE-TRAINED VS SPARSE ID-BASED EMBEDDINGS

|  | Pre-trained content embeddings | On-the-fly ID-based embeddings |
|---|---|---|
| Pros | World knowledge can help with generalization and cold-start, and embeddings are generic and can be applied to many downstream tasks. | Learning task-specific data patterns -> exploit prediction performance by memorizing e.g. co-purchase behaviors. |
| Cons | May not be suitable for exploitation tasks, e.g. modeling consumer repurchase patterns. | Risk of overfitting under limited training data or imbalanced data distribution. |

Table 5: The trade-off between pre-trained and on-the-fly embeddings.

## F  THE SELECTION OF PRE-TRAINED EMBEDDING DIMENSION

|  | Embedding size | Average MTEB [27] score | Pages per dollar |
|---|---|---|---|
| **text-embedding-ada-002** | 1536 | 61.0 | 12500 |
|  | 512 | 61.6 | 12500 |
| **text-embedding-3-small** | 1536 | 62.3 | 62500 |
|  | 256 | 62.0 | 62500 |
| **text-embedding-3-large** | 3072 | 64.6 | 9615 |
|  | 1024 | 64.1 | 9615 |

Table 6: OpenAI text-to-embedding model performance and cost benchmark. Pages-per-dollar is assuming ~800 tokens per page [2].